\documentclass[12pt,twoside]{article}
\def\,{\mskip 3mu} \def\>{\mskip 4mu plus 2mu minus 4mu} \def\;{\mskip 5mu plus 5mu} \def\!{\mskip-3mu}
\def\dispmuskip{\thinmuskip= 3mu plus 0mu minus 2mu \medmuskip=  4mu plus 2mu minus 2mu \thickmuskip=5mu plus 5mu minus 2mu}
\def\textmuskip{\thinmuskip= 0mu                    \medmuskip=  1mu plus 1mu minus 1mu \thickmuskip=2mu plus 3mu minus 1mu}
\textmuskip
\def\beq{\dispmuskip\begin{equation}}    \def\eeq{\end{equation}\textmuskip}
\def\beqn{\dispmuskip\begin{displaymath}}\def\eeqn{\end{displaymath}\textmuskip}
\def\bqa{\dispmuskip\begin{eqnarray}}    \def\eqa{\end{eqnarray}\textmuskip}
\def\bqan{\dispmuskip\begin{eqnarray*}}  \def\eqan{\end{eqnarray*}\textmuskip}

\newtheorem{theorem}{Theorem}
\newtheorem{corollary}[theorem]{Corollary}
\newtheorem{lemma}[theorem]{Lemma}
\newtheorem{definition}[theorem]{Definition}

\newenvironment{keywords}{\centerline{\bf\small
Keywords}\vspace{-1ex}\begin{quote}\small}{\par\end{quote}\vskip
1ex}
\newtheorem{tablex}[theorem]{Table}
\newtheorem{figurex}[equation]{Figure}

\def\ftheorem#1#2#3{\begin{theorem}[#2]\label{#1} #3 \end{theorem} }

\def\fdefinition#1#2#3{\begin{definition}[#2]\label{#1} #3 \end{definition} }
\def\ftablex#1#2#3{\begin{tablex}[#2]\label{#1} #3 \end{tablex} }

\def\idx#1{\index{#1}#1} 
\def\indxs#1#2{\index{#1!#2}\index{#2!#1}} 
\def\paragraph#1{\vspace{1ex}\noindent{\bf{#1.}}}

\def\toinfty#1{\stackrel{#1\to\infty}{\longrightarrow}}
\def\nq{\hspace{-1em}}

\def\odt{{\textstyle{1\over 2}}}
\def\odf{{\textstyle{1\over 4}}}
\def\eps{\varepsilon}                   
\def\epstr{\epsilon}                    
\def\qmbox#1{{\quad\mbox{#1}\quad}}

\def\eqm{\stackrel\times=}             

\def\geqm{\stackrel\times\geq}

\def\l{{l}}                             
\def\M{{\cal M}}                        
\def\X{{\cal X}}                        

\def\E{{\bf E}}                         
\def\P{{\bf P}}                         
\def\B{\{0,1\}}                        
\def\MM{M}                              
\def\th{\theta}

\def\Set#1{{\if#1Q{I\!\!\!#1}\else\if#1Z{Z\!\!\!Z}\else{I\!\!#1}\fi\fi}}

\def\sumprime{\mathop{{\sum\nolimits'}}}

\begin{document}

\title{\vskip -15mm\normalsize\sc Technical Report \hfill IDSIA-05-03
\vskip 2mm\bf\LARGE\hrule height5pt \vskip 3mm
\sc On the Existence and Convergence \\ of Computable Universal Priors\thanks{%
This work was supported by SNF grant 2000-61847.00 to J\"{u}rgen
Schmidhuber.}
\vskip 6mm \hrule height2pt \vskip 5mm}
\author{{\bf Marcus Hutter}\\[3mm]
\normalsize IDSIA, Galleria 2, CH-6928\ Manno-Lugano, Switzerland\\
\normalsize marcus@idsia.ch \hspace{8.5ex} http://www.idsia.ch/$^{_{_\sim}}\!$marcus}
\date{29 May 2003}
\maketitle

\begin{abstract}
\noindent Solomonoff unified Occam's razor and Epicurus' principle
of multiple explanations to one elegant, formal, universal theory
of inductive inference, which initiated the field of algorithmic
information theory. His central result is that the posterior of
his universal semimeasure $\MM$ converges rapidly to the true
sequence generating posterior $\mu$, if the latter is computable.
Hence, $M$ is eligible as a universal predictor in case of unknown
$\mu$.
We investigate the existence and convergence of computable
universal (semi)measures for a hierarchy of computability classes:
finitely computable, estimable, enumerable, and approximable.
For instance, $\MM$ is known to be enumerable, but not finitely
computable, and to dominate all enumerable semimeasures.
We define seven classes of (semi)measures based on these four
computability concepts. Each class may or may not contain a
(semi)measure which dominates all elements of another class. The
analysis of these 49 cases can be reduced to four basic cases, two
of them being new. The results hold for discrete and continuous
semimeasures.
We also investigate more closely the types of convergence, possibly
implied by universality: in difference and in ratio, with probability
1, in mean sum, and for Martin-L{\"o}f random sequences.
We introduce a generalized concept of randomness for individual
sequences and use it to exhibit difficulties regarding these
issues.
\end{abstract}


\pagebreak
\section{Introduction}\label{secIntro}

All induction problems can be phrased as sequence prediction
tasks. This is, for instance, obvious for time series prediction,
but also includes classification tasks. Having observed data $x_t$
at times $t<n$, the task is to predict the $t$-th symbol $x_t$
from sequence $x=x_1...x_{t-1}$.
The key concept to attack general induction problems is
Occam's razor and to a less extend Epicurus' principle of
multiple explanations. The former/latter may be interpreted as to
keep the simplest/all theories consistent with the observations
$x_1...x_{t-1}$ and to use these theories to predict $x_t$.
Solomonoff \cite{Solomonoff:64,Solomonoff:78} formalized and
combined both principles in his universal prior $\MM(x)$ which
assigns high/low probability to simple/complex environments, hence
implementing Occam and Epicurus.
Solomonoff's \cite{Solomonoff:78} central result is that if
the probability $\mu(x_t|x_1...x_{t-1})$ of observing $x_t$ at
time $t$, given past observations $x_1...x_{t-1}$ is
a computable function, then the
universal posterior
$\MM(x_t|x_1...x_{t-1})$
converges rapidly for $t\to\infty$ to the true posterior
$\mu(x_t|x_1...x_{t-1})$, hence
$\MM$ represents a universal predictor in case of unknown $\mu$.

One representation of $\MM$ is as a weighted sum of
{\em all} enumerable ``defective'' probability measures, called
semimeasures (see Definition \ref{defSemi}).
The (from this representation obvious) dominance $\MM(x)\geq
const.\times\mu(x)$ for all computable $\mu$ is the central
ingredient in the convergence proof.
%
What is so special about the class of all enumerable
semimeasures $\M_{enum}^{semi}$? The larger we choose $\M$ the
less restrictive is the essential assumption that $\M$ should
contain the true distribution $\mu$.
Why not restrict to the still rather general class of estimable or
finitely computable (semi)measures? For {\em every} countable
class $\M$ and $\xi_\M(x):=\sum_{\nu\in\M} w_\nu \nu(x)$ with
$w_\nu>0$, the important dominance $\xi_\M(x)\geq w_\nu
\nu(x)\,\forall\nu\in\M$ is satisfied. The question is what
properties does $\xi_\M$ possess. The distinguishing property of
$\MM=\xi_{\M_{enum}^{semi}}$ is that it is itself
an element of $\M_{enum}^{semi}$.
On the other hand, for prediction $\xi_\M\in\M$ is not by itself
an important property. What matters is  whether $\xi_\M$ is
computable (in one of the senses defined) to avoid
getting into the (un)realm of non-constructive math.

The intention of this work is to investigate the existence,
computability and convergence of universal (semi)measures for
various computability classes: finitely computable $\subset$
estimable $\subset$ enumerable $\subset$ approximable (see
Definition \ref{defCompFunc}). For instance, $\MM(x)$ is
enumerable, but not finitely computable. The research in this work
was motivated by recent generalizations of Kolmogorov complexity
and Solomonoff's prior by Schmidhuber \cite{Schmidhuber:02gtm} to
approximable (and others not here discussed) cases.

\paragraph{Contents}
In Section \ref{secCC} we review various computability concepts
and discuss their relation.
In Section \ref{secUniM} we define the prefix Kolmogorov
complexity $K$, the concept of (semi)measures, Solomonoff's
universal prior $\MM$, and explain its universality.
Section \ref{secUSP} summarizes Solomonoff's major convergence
result, discusses general mixture distributions and the important
universality property -- multiplicative dominance.
In Section \ref{secUSM} we define seven classes of (semi)measures
based on four computability concepts. Each class may or may not
contain a (semi)measures which dominates all elements of another
class. We reduce the analysis of these 49 cases to four basic
cases. Domination (essentially by $\MM$) is known to be true for
two cases. The two new cases do not allow for domination.
In Section \ref{secConv} we investigate more closely the type of
convergence implied by universality. We summarize the result on
posterior convergence in difference $(\xi-\mu\to 0)$ and improve
the previous result \cite{Li:97} on the convergence in ratio
$\xi/\mu\to 1$ by showing rapid convergence without use
of Martingales.
In Section \ref{secMLconv} we investigate whether convergence for
all Martin-L{\"o}f random sequences could hold. We define a
generalized concept of randomness for individual sequences and use
it to show that proofs based on universality cannot decide this
question.
Section \ref{secConc} concludes the paper. Proofs will be
presented elsewhere.

\paragraph{Notation}
We denote strings of length $n$ over finite alphabet $\X$ by
$x=x_1x_2...x_n$ with $x_t\in\X$ and further abbreviate
$x_{1:n}:=x_1x_2...x_{n-1}x_n$ and $x_{<n}:=x_1... x_{n-1}$,
$\epstr$ for the empty string, $\l(x)$ for the length of string $x$,
and $\omega=x_{1:\infty}$ for infinite sequences.
%
We abbreviate $\lim_{n\to\infty}[f(n)-g(n)]=0$ by
$f(n)\toinfty{n}g(n)$ and say $f$ converges to $g$, without
implying that $\lim_{n\to\infty}g(n)$ itself exists. We write
$f(x)\geqm  g(x)$ for $g(x)=O(f(x))$.

\section{Computability Concepts}\label{secCC}
We define several computability concepts weaker than can be captured
by halting Turing machines.

\fdefinition{defCompFunc}{Computable functions}{
We consider functions $f:\Set{N}\to\Set{R}$:
\begin{itemize}
\item[]
$\nq f$ is {\em finitely computable} or {\em recursive} {\it iff}
there are Turing machines $T_{1/2}$ with output interpreted as natural
numbers and $f(x)={T_1(x)\over T_2(x)}$,
\item[]
$\nq f$ is {\em approximable} {\it iff}
$\phi(\cdot,\cdot)$ is finitely computable and
$\lim_{t\to\infty}\phi(x,t)=f(x)$.
\item[]
$\nq f$ is {\em lower semi-computable} or {\em enumerable} {\it
iff} additionally $\phi(x,t)\leq\phi(x,t+1)$.
\item[]
$\nq f$ is {\em upper semi-computable} or {\em co-enumerable} {\it
iff} $[-f]$ is lower semi-computable.
\item[]
$\nq f$ is {\em semi-computable} {\it iff} $f$ is lower- {\it or}
upper semi-computable.
\item[]
$\nq f$ is {\em estimable} {\it iff} $f$ is lower- {\it and} upper
semi-computable.
\end{itemize}
}

\noindent If $f$ is estimable we can finitely compute an
$\eps$-approximation of $f$ by upper and lower semi-computing $f$
and terminating when differing by less than $\eps$. This means
that there is a Turing machine which, given $x$ and $\eps$,
finitely computes $\hat y$ such that $|\hat y-f(x)|<\eps$.
Moreover it gives an interval estimate $f(x)\in[\hat y-\eps,\hat
y+\eps]$. An estimable integer-valued function is finitely
computable (take any $\eps<1$).
Note that if $f$ is only approximable or semi-computable we can
still come arbitrarily close to $f(x)$ but we cannot devise a
terminating algorithm which produces an $\eps$-approximation. In
the case of lower/upper semi-computability we can at least
finitely compute lower/upper bounds to $f(x)$. In case of
approximability, the weakest computability form, even this
capability is lost.
In analogy to lower/upper semi-computability one may think of
notions like lower/upper estimability but they are easily shown to
coincide with estimability. The following implications are valid:

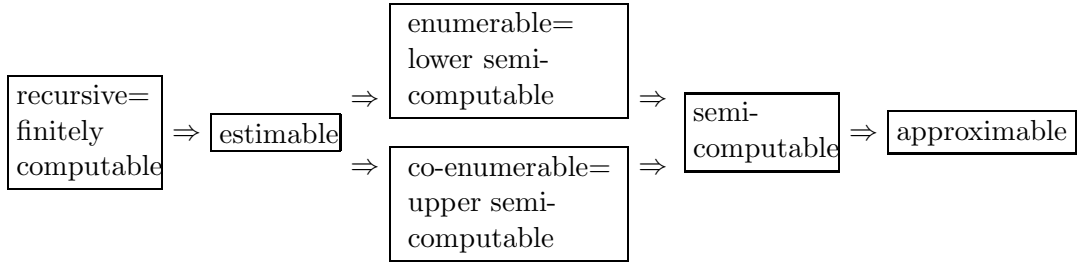
\begin{center}\small
\fbox{\parbox{11ex}{recursive=\\ finitely\\ computable}}
$\Rightarrow$
\fbox{\parbox{9ex}{estimable}}
\parbox{26ex}{\raisebox{-3ex}{$\Rightarrow$} \fbox{
\parbox{17ex}{enumerable=\\lower semi-\\ computable}}
\raisebox{-3ex}{$\Rightarrow$} \\[2ex]
\raisebox{3ex}{$\Rightarrow$} \fbox{
\parbox{17ex}{co-enumerable=\\ upper semi-\\
computable}} \raisebox{3ex}{$\Rightarrow$}}
\fbox{\parbox{11ex}{semi-\\ computable}}
$\Rightarrow$
\fbox{approximable}
\end{center}

\noindent In the following we use the term computable synonymous
to finitely computable, but sometimes also generically for some of
the computability forms of Definition \ref{defCompFunc}.
What we call {\em estimable} is often just called {\em
computable}, but it makes sense to separate the concepts of
finite computability and estimability in this work, since the
former is conceptually easier and some previous results have only
been proved for this case.

\section{The Universal Prior $\MM$}\label{secUniM}
\index{Turing machine!universal}
\index{Turing machine!prefix}
\index{tape!unidirectional}
\index{tape!bidirectional}%
\index{semimeasure!universal}
%
The prefix Kolmogorov complexity $K(x)$ is defined as the length
of the shortest binary program $p\in\B^*$ for which a universal prefix
Turing machine $U$ (with binary program tape and $\X$ary output
tape) outputs string $x\in\X^*$, and similarly $K(x|y)$ in case of side
information $y$ \cite{Li:97}:
\beqn
  K(x)=\min\{\l(p):U(p)=x\},\qquad
  K(x|y)=\min\{\l(p):U(p,y)=x\}
\eeqn
Solomonoff
\cite{Solomonoff:64,Solomonoff:78}
(with a flaw fixed by Levin \cite{Zvonkin:70})
defined (earlier) the closely related
quantity, the universal prior $\MM(x)$.
It is defined as the
probability that the output of a universal Turing machine starts
with $x$ when provided with \idx{fair coin flips} on the input
tape. Formally, $\MM$ can be defined as
\beq\label{Mdef}
  \MM(x)\;:=\;\sum_{p\;:\;U(p)=x*}\nq 2^{-\l(p)}
\eeq
where the sum is over all so called minimal programs $p$ for which
$U$ outputs a string starting with $x$ (indicated by the $*$).
Before we can discuss the stochastic properties of $\MM$ we
need the concept of (semi)measures for strings.

\index{semimeasure!enumerable}
\fdefinition{defSemi}{Continuous (Semi)measures}{
$\mu(x)$ denotes the probability that a sequence starts
with string $x$. We call $\mu\geq 0$ a (continuous) semimeasure if
$\mu(\epstr)\leq 1$ and $\mu(x)\geq\mu(x0)+\mu(x1)$, and a
(probability) measure if equality holds.
}

\noindent We have $\MM(x0)+\MM(x1)<\MM(x)$ because there are
programs $p$, which output $x$, neither followed by $0$ nor $1$.
They just stop after printing $x$ or continue forever without any
further output. Together with $\MM(\epstr)=1$ this shows that $\MM$
is a semimeasure, but {\it not} a probability measure. We can now
state the fundamental property of $\MM$ \cite{Solomonoff:78}:

\ftheorem{thUniM}{Universality of $\MM$}{
The universal prior $\MM$ is an enumerable semimeasure which
multiplicatively dominates all enumerable semimeasures in the
sense that $\MM(x) \;\geqm\; 2^{-K(\rho)}\cdot \rho(x)$
for all an enumerable semimeasures $\rho$. $\MM$ is enumerable, but not
estimable or finitely computable.
}
\indxs{multiplicative}{majorization}
\indxs{probability distribution}{computable}

\noindent The Kolmogorov complexity of a function like $\rho$ is
defined as the length of the shortest self-delimiting code of a
Turing machine computing this function in the sense of Definition
\ref{defCompFunc}. Up to a multiplicative constant, $\MM$ assigns higher
probability to all $x$ than any other computable probability
distribution.

It is possible to normalize $\MM$ to a true probability measure
$\MM_{norm}$ \cite{Solomonoff:78,Li:97} with dominance still being
true, but at the expense of giving up enumerability ($\MM_{norm}$
is still approximable). $\MM$ is more convenient when studying
algorithmic questions, but a true probability measure like
$\MM_{norm}$ is more convenient when studying stochastic questions.

\section{Universal Sequence Prediction}\label{secUSP}

In which sense does $\MM$ incorporate Occam's razor and Epicurus'
principle of multiple explanations? Since the shortest programs
$p$ dominate the sum in $M$, $\MM(x)$ is roughly equal to
$2^{-K(x)}$ ($\MM(x)=2^{-K(x)+O(K(\l(x))}$), i.e.\
$\MM$ assigns high probability to simple
strings. More useful is to think of $x$ as being the observed
history. We see from (\ref{Mdef}) that every program $p$
consistent with history $x$ is allowed to contribute to $\MM$
(Epicurus). On the other hand shorter programs give significantly
larger contribution (Occam). How does all this affect prediction?
If $\MM(x)$ describes our (subjective) prior belief in $x$, then
$\MM(y|x):=\MM(xy)/\MM(x)$ must be our posterior belief in $y$.
From the symmetry of algorithmic information $K(xy)\approx
K(y|x)+K(x)$, and $\MM(x)\approx 2^{-K(x)}$ and $\MM(xy)\approx
2^{-K(xy)}$ we get $\MM(y|x)\approx 2^{-K(y|x)}$. This tells us
that $\MM$ predicts $y$ with high probability iff $y$ has an easy
explanation, given $x$ (Occam \& Epicurus).

The above qualitative discussion should not create the impression
that $\MM(x)$ and $2^{-K(x)}$ always lead to predictors of
comparable quality. Indeed in the online/incremental setting,
$K(y)=O(1)$ invalidates the consideration above. The proof of
(\ref{eukdist}) below, for instance, depends on $\MM$ being a
semimeasure and the chain rule being exactly true, neither of them is
satisfied by $2^{-K(x)}$. See \cite{Hutter:03unimdl} for a more
detailed analysis.

\index{sequence prediction!Solomonoff} Sequence
prediction algorithms try to predict the continuation $x_t\in\X$
of a given sequence $x_1...x_{t-1}$.
We assume that the true sequence is
drawn from a computable
probability distribution $\mu$, i.e.\ the true (objective)
probability of $x_{1:t}$ is $\mu(x_{1:t})$. The probability of
$x_t$ given $x_{<t}$ hence is
$\mu(x_t|x_{<t})=\mu(x_{1:t})/\mu(x_{<t})$.
Solomonoff's \cite{Solomonoff:78} central result is that $\MM$
converges to $\mu$. More precisely, for binary alphabet, he showed that
\beq\label{eukdist}
  \sum_{t=1}^\infty
  \nq\nq\;\sum_{\qquad x_{<t}\in\B^{t-1}}\nq\nq\;
  \mu(x_{<t}) \Big(\MM(0|x_{<t})-\mu(0|x_{<t})\Big)^2
  \;\leq\;
  {\odt}\ln 2\!\cdot\!K(\mu)+O(1) \;<\; \infty.
\eeq
The infinite sum can only be finite if the difference
$\MM(0|x_{<t})-\mu(0|x_{<t})$ tends to zero for $t\to\infty$ with
$\mu$ probability $1$ (see Definition \ref{defConv}$(i)$ and
\cite{Hutter:01alpha} or Section \ref{secConv} for general
alphabet). This holds for {\it any} computable probability
distribution $\mu$. The reason for the astonishing property of a
single (universal) function to converge to {\it any} computable
probability distribution lies in the fact that the set of
$\mu$-random sequences differ for different $\mu$. Past data
$x_{<t}$ are exploited to get a (with $t\to\infty$) improving
estimate $\MM(x_t|x_{<t})$ of $\mu(x_t|x_{<t})$.

The universality property (Theorem \ref{thUniM}) is the central
ingredient in the proof of (\ref{eukdist}). The proof
involves the construction of a semimeasure $\xi$
whose dominance is obvious. The hard part is to show its
enumerability and equivalence to $\MM$.
Let $\M$ be the (countable) set of all enumerable semimeasures
and define
\beq\label{xidef}
  \xi(x):=\sum_{\nu\in\M}2^{-K(\nu)}\nu(x).
\eeq
Then dominance
\beq\label{xidom}
 \xi(x)\geq 2^{-K(\nu)}\nu(x)\quad\forall\,\nu\in\M
\eeq
is obvious. Is $\xi$ lower semi-computable? To answer this
question one has to be more precise. Levin \cite{Zvonkin:70} has
shown that the set of {\em all} lower semi-computable semimeasures
is enumerable (with repetitions). For this (ordered multi) set
$\M=\M_{enum}^{semi}:=\{\nu_1,\nu_2,\nu_3,...\}$ and
$K(\nu_i):=K(i)$ one can easily see that $\xi$ is lower
semi-computable. Finally proving $\MM(x)\eqm\xi(x)$ also
establishes universality of $\MM$ (see \cite{Solomonoff:78,Li:97}
for details).

The advantage of $\xi$ over $\MM$ is that it immediately
generalizes to arbitrary weighted sums of (semi)measures
for arbitrary countable $\M$.

\section{Universal (Semi)Measures}\label{secUSM}

What is so special about the set of all enumerable
semimeasures $\M_{enum}^{semi}$? The larger we choose $\M$ the less restrictive
is the assumption that $\M$ should contain the true distribution
$\mu$, which will be essential throughout the paper.
Why do not restrict to the still rather general class of estimable
or finitely computable (semi)measures? It is clear that for every
countable set $\M$,
\beq\label{defxi}
  \xi(x):=\xi_\M(x):=\sum_{\nu\in\M} w_\nu \nu(x)
  \qmbox{with} \sum_{\nu\in\M}w_\nu\leq 1 \qmbox{and} w_\nu>0
\eeq
dominates all $\nu\in\M$. This dominance is
necessary for the desired convergence $\xi\to\mu$ similarly to
(\ref{eukdist}). The question is what properties $\xi$ possesses.
The distinguishing property of $\M_{enum}^{semi}$ is that $\xi$ is
itself an element of $\M_{enum}^{semi}$. When concerned with
predictions, $\xi_\M\in\M$ is not by itself an important property,
but whether $\xi$ is computable in one of the senses of Definition
\ref{defCompFunc}. We define
\bqan
 \M_1\geqm\M_2 & :\Leftrightarrow &
 \mbox{there is an element of $\M_1$ which dominates all elements of
 $\M_2$} \\
 & :\Leftrightarrow &
\exists\rho\!\in\!\M_1\;\forall\nu\!\in\!\M_2\;\exists w_\nu\!>\!0
\;\forall x:\rho(x)\!\geq\!w_\nu\nu(x).
\eqan
$\geqm $ is transitive (but not necessarily reflexive) in the
sense that $\M_1 \geqm \M_2 \geqm \M_3$ implies $\M_1 \geqm \M_3$
and $\M_0 \supseteq \M_1 \geqm \M_2 \supseteq \M_3$ implies $\M_0
\geqm \M_3$.
For the computability concepts introduced in Section \ref{secCC}
we have the following proper set inclusions
\beqn
\begin{array}{ccccccc}
  \M_{comp}^{msr}  & \subset & \M_{est}^{msr}  & \equiv  & \M_{enum}^{msr}  & \subset & \M_{appr}^{msr} \\
        \cap       &         &      \cap       &         &       \cap       &         &     \cap        \\
  \M_{comp}^{semi} & \subset & \M_{est}^{semi} & \subset & \M_{enum}^{semi} & \subset & \M_{appr}^{semi}
\end{array}
\eeqn
where $\M^{msr}_c$ stands for the set of all probability measures
of appropriate computability type $c\in\{$comp=finitely
computable, est=estimable, enum=enumerable,
appr=approximable$\}$, and similarly for semimeasures
$\M^{semi}_c$. From an enumeration of a measures $\rho$ on can
construct a co-enumeration by exploiting
$\rho(x_{1:n})=1-\sum_{y_{1:n}\neq x_{1:n}}\rho(y_{1:n})$. This
shows that every enumerable measure is also co-enumerable, hence
estimable, which proves the identity $\equiv$ above.

With this notation, Theorem \ref{thUniM} implies
$\M_{enum}^{semi}\geqm\M_{enum}^{semi}$. Transitivity allows to
conclude, for instance, that
$\M_{appr}^{semi}\geqm\M_{comp}^{msr}$, i.e.\ that there is an
approximable semimeasure which dominates all computable measures.

The standard ``diagonalization'' way of proving
$\M_1\stackrel\times{\not\geq}\M_2$ is to take an arbitrary
$\mu\in\M_1$ and ``increase'' it to $\rho$ such that
$\mu\stackrel\times{\not\geq}\rho$ and show that $\rho\in\M_2$.
There are $7\times 7$ combinations of (semi)measures $\M_1$ with
$\M_2$ for which $\M_1\geqm\M_2$ could be true or false. There are
four basic cases, explicated in the following theorem, from which
the other 49 combinations displayed in Table \ref{tabUniSMsr}
follow by transitivity.

\ftheorem{thNoUniApp}{Universal (semi)measures}{
A semimeasure $\rho$ is said to be universal for $\M$ if it
multiplicatively dominates all elements of $\M$ in the sense
$\forall\nu\exists w_\nu>0:\rho(x)\geq w_\nu\nu(x)\forall x$. The
following holds true:
\begin{itemize}
\item[$o)$]
$\exists\rho:\{\rho\}\geqm\M$: For every countable set
of (semi)measures $\M$, there is a (semi)measure which dominates
all elements of $\M$.
\item[$i)$]
$\M_{enum}^{semi}\geqm\M_{enum}^{semi}$:
The class of enumerable semimeasures {\em contains}
a universal element.
\item[$ii)$]
$\M_{appr}^{msr}\geqm\M_{enum}^{semi}$:
There {\em is} an approximable measure which dominates all enumerable
semimeasures.
\item[$iii)$]
$\M_{est}^{semi}\stackrel\times{\not\geq}\M_{comp}^{msr}$: There is
{\em no} estimable semimeasure which dominates all computable
measures.
\item[$iv)$]
$\M_{appr}^{semi}\stackrel\times{\not\geq}\M_{appr}^{msr}$: There is
{\em no} approximable semimeasure which dominates all approximable
measures.
\end{itemize}
}

\begin{table}[thb]
\ftablex{tabUniSMsr}{Existence of universal (semi)measures}{%
The entry in row $r$ and column $c$ indicates whether there is a
$r$-able (semi)measure $\rho$ for the set $\M$ which contains all
$c$-able (semi)measures, where $r,c\in\{$comput, estimat, enumer,
approxim$\}$. Enumerable measures are estimable. This is the
reason why the enum. row and column in case of measures is
missing. The superscript indicates from which part of Theorem
\ref{thNoUniApp} the answer follows. For the bold face entries
directly, for the others using transitivity of $\geqm $.
\begin{center}
\begin{tabular}{|c|c||c|c|c|c||c|c|c|}\hline
      $\nwarrow$ &  $\M$ & \multicolumn{4}{c||}{semimeasure} & \multicolumn{3}{c|}{measure}\\ \hline
$\rho$&$\searrow$& comp.      & est.       & enum.         & appr.     & comp.         & est.       & appr.        \\ \hline\hline
      s  & comp. & no$^{iii}$ & no$^{iii}$ & no$^{iii}$    & no$^{iv}$ & no$^{iii}$    & no$^{iii}$ & no$^{iv}$    \\ \cline{2-9}
      e  & est.  & no$^{iii}$ & no$^{iii}$ & no$^{iii}$    & no$^{iv}$ & {\bf no}$^{\bf iii}$& no$^{iii}$ & no$^{iv}$    \\ \cline{2-9}
      m  & enum. & yes$^{i}$  & yes$^{i}$  & {\bf yes}$^{\bf i}$ & no$^{iv}$ & yes$^{i}$     & yes$^{i}$  & no$^{iv}$    \\ \cline{2-9}
      i  &appr.  & yes$^{i}$  & yes$^{i}$  & yes$^{i}$     & no$^{iv}$ & yes$^{i}$     & yes$^{i}$  & {\bf no}$^{\bf iv}$\\ \hline\hline
      m  & comp. & no$^{iii}$ & no$^{iii}$ & no$^{iii}$    & no$^{iv}$ & no$^{iii}$    & no$^{iii}$ & no$^{iv}$    \\ \cline{2-9}
      s  & est.  & no$^{iii}$ & no$^{iii}$ & no$^{iii}$    & no$^{iv}$ & no$^{iii}$    & no$^{iii}$ & no$^{iv}$    \\ \cline{2-9}
      r  &appr.  & yes$^{ii}$ & yes$^{ii}$ & {\bf yes}$^{\bf ii}$& no$^{iv}$ & yes$^{ii}$    & yes$^{ii}$ & no$^{iv}$    \\ \hline
\end{tabular}
\end{center}}
\end{table}

\noindent If we ask for a universal (semi)measure which at least satisfies
the weakest form of computability, namely being approximable, we
see that the largest dominated set among the 7 sets defined above
is the set of enumerable semimeasures. This is the reason why
$\M_{enum}^{semi}$ plays a special role.
On the other hand, $\M_{enum}^{semi}$ is not the largest set
dominated by an approximable semimeasure, and indeed no such
largest set exists. One may, hence, ask for ``natural'' larger
sets $\M$. One such set, namely the set of cumulatively enumerable
semimeasures $\M_{CEM}$, has recently been discovered by
Schmidhuber \cite{Schmidhuber:02gtm}, for which even
$\xi_{CEM}\in\M_{CEM}$ holds.

\noindent Theorem \ref{thNoUniApp} also holds for {\em discrete
(semi)measures} $P$ defined as follows:

\index{semimeasure!enumerable}
\fdefinition{defDSemi}{Discrete (Semi)measures}{
$P(x)$ denotes the probability of $x\in\Set N$. We call
$P:\Set{N}\to[0,1]$ a discrete (semi)measure if $\sum_{x\in\Set{N}}
P(x)\stackrel{(<)}=1$.
}

\noindent Theorem \ref{thNoUniApp}
$(i)$ is Levin's major result \cite[Th.4.3.1 \& Th.4.5.1]{Li:97}, %
$(ii)$ is due to Solomonoff \cite{Solomonoff:78}, %
the proof of
$\M_{comp}^{semi}\stackrel\times{\not\geq}\M_{comp}^{semi}$ in
\cite[p249]{Li:97} contains minor errors and is not extensible to
$(iii)$ and the proof in \cite[p276]{Li:97} only applies to
infinite alphabet and not to the binary/finite case considered
here. A complete proof of $(o)-(iv)$ for discrete and continuous
(semi)measures is given elsewhere.

\section{Posterior Convergence}\label{secConv}

We have investigated in detail the computational properties of
various mixture distributions $\xi$. A mixture $\xi_\M$
multiplicatively dominates all distributions in $\M$. We have
mentioned that dominance implies posterior convergence. In this
section we present in more detail what dominance implies and what
not.

Convergence of $\xi(x_t|x_{<t})$ to $\mu(x_t|x_{<t})$ with
$\mu$-probability 1 tells us that $\xi(x_t|x_{<t})$ is close to
$\mu(x_t|x_{<t})$ for sufficiently large $t$ and ``most''
sequences $x_{1:\infty}$. It says nothing about the speed of
convergence, nor whether convergence is true for any {\em particular}
sequence (of measure 0). Convergence {\em in mean sum} defined
below is intended to capture the rate of convergence,
Martin-L\"{o}f randomness is used to capture convergence
properties for individual sequences.

Martin-L\"{o}f randomness is a very important concept of
randomness of individual sequences, which is closely related to
Kolmogorov complexity and Solomonoff's universal prior. Levin gave
a characterization equivalent to Martin-L\"{o}f's original
definition \cite{Levin:73random}:

\ftheorem{defML}{Martin-L\"{o}f random sequences}{
A sequence $x_{1:\infty}$ is $\mu$-Martin-L\"{o}f random
($\mu$.M.L.) iff there is a constant $c$ such that
$\MM(x_{1:n})\leq c\cdot \mu(x_{1:n})$ for all $n$.
}

\noindent
One can show that a $\mu$.M.L.\ random sequence $x_{1:\infty}$
passes {\em all} thinkable effective randomness tests, e.g.\ the
law of large numbers, the law of the iterated logarithm, etc.
In particular, the set of all $\mu$.M.L. random sequences has
$\mu$-measure 1.
The following generalization is natural when considering general
Bayes-mixtures $\xi$ as in this work:

\fdefinition{defmuMr}{$\mu/\xi$-random sequences}{
A sequence $x_{1:\infty}$ is called $\mu/\xi$-random
($\mu.\xi$.r.) iff there is a constant $c$ such that
$\xi(x_{1:n})\leq c\cdot \mu(x_{1:n})$ for all $n$.
}

Typically, $\xi$ is a mixture over some $\M$ as defined in
(\ref{xidef}), in which case the reverse inequality
$\xi(x)\geqm\mu(x)$ is also true (for all $x$). For finite $\M$ or
if $\xi\in\M$, the definition of $\mu/\xi$-randomness depends only
on $\M$, and not on the specific weights used in $\xi$. For
$\M=\M_{enum}^{semi}$, $\mu/\xi$-randomness is just $\mu$.M.L.\
randomness. The larger $\M$, the more patterns are recognized as
non-random.
Roughly speaking, those regularities characterized by some
$\nu\in\M$ are recognized by $\mu/\xi$-randomness, i.e.\ for
$\M\subset\M_{enum}^{semi}$ some $\mu/\xi$-random strings may not
be M.L.\ random.
Other randomness concepts, e.g.\ those by Schnorr, Ko, van
Lambalgen, Lutz, Kurtz, von Mises, Wald, and Church (see
\cite{Wang:96,Lambalgen:87,Schnorr:71}), could possibly also be
characterized in terms of $\mu/\xi$-randomness for particular
choices of $\cal M$.

\indxs{random sequence}{convergence} A classical (non-random)
real-valued sequence $a_t$ is defined to converge to $a_*$, short
$a_t\to a_*$ if $\forall\eps\exists t_0\forall t\geq
t_0:|a_t-a_*|<\eps$. We are interested in convergence properties
of random sequences $z_t(\omega)$ for $t\to\infty$ (e.g.\
$z_t(\omega)=\xi(\omega_t|\omega_{<t})-\mu(\omega_t|\omega_{<t})$).
We denote $\mu$-expectations by $\E$. The expected value of a
function $f:\X^t\to\Set R$, dependent on $x_{1:t}$, independent of
$x_{t+1:\infty}$, and possibly undefined on a set of $\mu$-measure
0, is $\E[f] =
\sumprime_{\!x_{1:t}\in\X^t}\mu(x_{1:t})f(x_{1:t})$. The prime
denotes that the sum is restricted to $x_{1:t}$ with
$\mu(x_{1:t})\neq 0$. Similarly we use $\P[..]$ to denote the
$\mu$-probability of event $[..]$
We define four convergence concepts for random sequences.

\index{convergence!with probability 1}%
\index{convergence!in the mean}
\index{convergence!in mean sum}
\index{convergence!in probability}
\index{convergence!Martin-L\"of}
\index{convergence!$\M$}
\fdefinition{defConv}{Convergence of random sequences}{
Let $z_1(\omega),z_2(\omega),...$ be a sequence of real-valued
random variables. $z_t$ is said to
converge for $t\to\infty$ to random variable $z_*(\omega)$
\begin{itemize}\itemindent8ex
\item[$i)$] with probability 1 (w.p.1) $:\Leftrightarrow$
  $\P[\{\omega:z_t\to z_*\}]=1$,
\item[$ii)$] in mean sum (i.m.s.) $:\Leftrightarrow$
$\sum_{t=1}^\infty\E[(z_t-z_*)^2]<\infty$,
\item[$iii)$] for every $\mu$-Martin-L{\"o}f random sequence ($\mu$.M.L.) $:\Leftrightarrow$ \\
\hspace*{8ex}$\forall\omega:$ $[\exists c\forall n:
\MM(\omega_{1:n})\leq c\mu(\omega_{1:n})]$
  implies $z_t(\omega)\to z_*(\omega)$ for $t\to\infty$,
\item[$iv)$] for every $\mu/\xi$-random sequence ($\mu.\xi$.r.) $:\Leftrightarrow$ \\
\hspace*{8ex}$\forall\omega:$ $[\exists c\forall n:
\xi(\omega_{1:n})\leq c\mu(\omega_{1:n})]$
  implies $z_t(\omega)\to z_*(\omega)$ for $t\to\infty$.
\end{itemize}
}

\noindent In statistics, $(i)$ is the ``default'' characterization of
convergence of random sequences.
Convergence i.m.s.\ $(ii)$ is very strong: it
provides a rate of convergence in the sense that the expected
number of times $t$ in which $z_t$ deviates more than $\eps$ from
$z_*$ is finite and bounded by
$\sum_{t=1}^\infty\E[(z_t-z_*)^2]/\eps^2$.
Nothing can be said for {\em which} $t$ these deviations occur.
If, additionally, $|z_t-z_*|$ were monotone decreasing, then
$|z_t-z_*|=o(t^{-1/2})$ could be concluded.
$(iii)$ uses Martin-L\"{o}f's notion of randomness of {\em individual}
sequences to define convergence M.L. Since this work
deals with general Bayes-mixtures $\xi$, we generalized in $(iv)$
the definition of convergence M.L.\ based on $\MM$ to
convergence $\mu.\xi$.r.\ based on $\xi$ in a natural way.
One can show that convergence i.m.s.\ implies convergence w.p.1.
Also convergence M.L.\ implies convergence w.p.1.
\index{random sequence!convergence relations}
\index{convergence!relations}
Universality of $\xi$ implies the following posterior convergence results:

\index{convergence!$\xi$ to $\mu$}

\ftheorem{thConv}{Convergence of $\xi$ to $\mu$}{
Let there be sequences $x_1x_2...$ over a finite alphabet $\X$
drawn with probability $\mu(x_{1:n})\in\M$ for the first $n$
symbols, where $\mu$ is a measure. The universal posterior
probability $\xi(x_t|x_{<t})$
of the next symbol $x_t$ given $x_{<t}$ 
is related to the true posterior probability $\mu(x_t|x_{<t})$
in the following way:\vspace{-1ex}
\beqn
   \sum_{t=1}^n\E{\textstyle\left[\left(\sqrt{{\xi(x_t|x_{<t})
          \over\mu(x_t|x_{<t})}}-1\right)^2\right]} \;\leq\;
   \sum_{t=1}^n\E\bigg[\sum_{x'_t}
        \left(\sqrt{\xi(x'_t|x_{<t})}-\sqrt{\mu(x'_t|x_{<t})}\right)^2\bigg]
        \;\leq\; \ln{w_\mu^{-1}} \;<\; \infty
\eeqn
where $w_\mu$ is the weight (\ref{defxi}) of $\mu$ in $\xi$.
}

\noindent Theorem \ref{thConv} implies
\beqn
 \mbox{$\sqrt{\xi(x'_t|x_{<t})} \to \sqrt{\mu(x'_t|x_{<t})}$
 for any $x'_t$ and
 $\sqrt{{\xi(x_t|x_{<t})\over\mu(x_t|x_{<t})}} \to 1$, both
 i.m.s.\ for $t\to\infty$}.
\eeqn
%
\indxs{semi-martingale}{convergence}\index{martingales}%
\noindent The latter strengthens the result
$\xi(x_t|x_{<t})/\mu(x_t|x_{<t})\to 1$ w.p.1 derived by G\'acs in
\cite[Th.5.2.2]{Li:97} in that it also provides the ``speed'' of
convergence.

Note also the subtle difference between the two convergence
results. For {\em any} sequence $x'_{1:\infty}$ (possibly constant
and not necessarily $\mu$-random),
$\mu(x'_t|x_{<t})-\xi(x'_t|x_{<t})$ converges to zero w.p.1
(referring to $x_{1:\infty}$), but no statement is possible for
$\xi(x'_t|x_{<t})/\mu(x'_t|x_{<t})$, since
$\lim\,\inf\mu(x'_t|x_{<t})$ could be zero. On the other hand, if
we stay {\em on} the $\mu$-random sequence ($x'_{1:\infty} =
x_{1:\infty}$), we have $\xi(x_t|x_{<t})/\mu(x_t|x_{<t})
\to 1$ (whether $\inf\mu(x_t|x_{<t})$ tends to zero or not does
not matter).
Indeed, it is easy to see that $\xi(1|0_{<t})/\mu(1|0_{<t})\propto
t\to\infty$ diverges for $\M=\{\mu,\nu\}$, $\mu(1|x_{<t}):=\odt
t^{-3}$ and $\nu(1|x_{<t}):=\odt t^{-2}$, although $0_{1:\infty}$ is
$\mu$-random. 
%

\section{Convergence in Martin-L{\"o}f Sense}\label{secMLconv}

An interesting open question is whether $\xi$ converges to $\mu$
(in difference or ratio) individually for all Martin-L\"{o}f
random sequences. Clearly, convergence $\mu$.M.L. may at most fail
for a set of sequences with $\mu$-measure zero. A convergence
M.L.\ result would be particularly interesting and natural for
Solomonoff's universal prior $M$, since M.L.\ randomness can be
defined in terms of $\MM$ (see Theorem \ref{defML}). Attempts to
convert the bounds in Theorem \ref{thConv} to effective
$\mu$.M.L.\ randomness tests fail, since $M(x_t|x_{<t})$ is not
enumerable. The proof given of $M/\mu\stackrel{M.L.}\longrightarrow 1$
in \cite[Th.5.2.2]{Li:97} and \cite[Th.10]{Vitanyi:00} is
incomplete.$\!$\footnote{The formulation of their Theorem is quite
misleading in general: ``{\it Let $\mu$ be a positive recursive
measure. If the length of $y$ is fixed and the length of $x$ grows
to infinity, then $M(y|x)/\mu(y|x)\to 1$ with $\mu$-probability
one. The infinite sequences $\omega$ with prefixes $x$ satisfying
the displayed asymptotics are precisely [`$\Rightarrow$' {\em and}
`$\Leftarrow$'] the $\mu$-random sequences.}'' First, for
off-sequence $y$ convergence w.p.1 does not hold ($xy$ must be
demanded to be a prefix of $\omega$). Second, the proof of
`$\Leftarrow$' is loopy (see main text). Last, `$\Rightarrow$' is
given without proof and is probably wrong. Also the assertion in
\cite[Th.5.2.1]{Li:97} that $S_t:=\E\sum_{x'_t}
(\mu(x'_t|x_{<t})-M(x'_t|x_{<t}))^2$ converges to zero faster than
$1/t$ cannot be made, since $S_t$ may not decrease monotonically.}
The implication ``$\MM(x_{1:n})\leq c\cdot\mu(x_{1:n})\forall
n\Rightarrow \lim_{n\to\infty}\MM(x_{1:n})/\mu(x_{1:n})$ exists''
has been used, but not proven, and may indeed be wrong.

Vovk \cite{Vovk:87} shows that for two finitely computable
semi-measures $\mu$ and $\rho$ and $x_{1:\infty}$ being $\mu$
{\em and} $\rho$ M.L.\ random that
\beqn
\sum_{t=1}^\infty\sum_{x'_t}\left(\sqrt{\mu(x'_t|x_{<t})}-\sqrt{\rho(x'_t|x_{<t})}\right)^2<\infty
\qmbox{and}
\sum_{t=1}^\infty\left({\rho(x_t|x_{<t})\over\mu(x_t|x_{<t})}-1\right)^2<\infty.
\eeqn
If $\MM$ were recursive, then this would imply posterior
$\MM\to\mu$ and $\MM/\mu\to 1$ for every $\mu$.M.L.\ random
sequence $x_{1:\infty}$, since {\em every} sequence is $\MM$.M.L.\
random. Since $\MM$ is {\em not} recursive Vovk's theorem cannot
be applied and it is not obvious how to generalize it. So the
question of individual convergence remains open. More generally,
one may ask whether $\xi_\M\to\mu$ for every $\mu/\xi$-random
sequence. It turns out that this is true for some $\M$, but false for others.

\ftheorem{thMLConv}{$\mu/\xi$-convergence of $\xi$ to $\mu$}{
Let $\X=\B$ be binary and
$\M_\Theta:=\{\mu_\th:\mu_\th(1|x_{<t})=\th\,\forall t,\;
\th\in\Theta\}$ be the set of Bernoulli($\th$) distributions
with parameters $\th\in\Theta$. Let $\Theta_D$ be a countable
dense subset of $[0,1]$, e.g.\ $[0,1]\cap\Set Q$ and let $\Theta_G$
be a countable subset of $[0,1]$ with a gap in the sense that
there exist $0<\th_0<\th_1<1$ such that
$[\th_0,\th_1]\cap\Theta_G=\{\th_0,\th_1\}$, e.g.\
$\Theta_G=\{\odf,\odt\}$ or $\Theta_G=([0,{1\over
4}]\cup[{1\over 2},1])\cap\Set Q$. Then
\begin{itemize}
\item[$i)$] If $x_{1:\infty}$ is $\mu/\xi_{\M_{\Theta_D}}$ random with
$\mu\in\M_{\Theta_D}$, then $\xi_{\M_{\Theta_D}}(x_t|x_{<t})\to\mu(x_t|x_{<t})$,
\item[$ii)$] There are $\mu\in\M_{\Theta_G}$ and $\mu/\xi_{\M_{\Theta_G}}\!\!$
random $x_{1:\infty}$ for which
$\xi_{\M_{\Theta_G}}\!\!(x_t|x_{<t})\not\to\mu(x_t|x_{<t})\!\!$
\end{itemize}\vspace{-1ex}
}

\noindent Our original/main motivation of studying
$\mu/\xi$-randomness is the implication of Theorem \ref{thMLConv}
that $\MM\stackrel{\mbox{\tiny M.L.}}\longrightarrow\mu$ cannot be
decided from $M$ being a mixture distribution or from the
universality property (Theorem \ref{thUniM}) alone. Further
structural properties of $\M_{enum}^{semi}$ have to be employed.
For Bernoulli sequences, convergence $\mu.\xi_{\M_\Theta}$.r.\ is
related to denseness of $\M_\Theta$. Maybe a denseness
characterization of $\M_{enum}^{semi}$ can solve the question of
convergence M.L.\ of $M$. The property $\MM\in\M_{enum}^{semi}$ is
also not sufficient to resolve this question, since there are
$\M\ni\xi$ for which $\xi\stackrel{\mu.\xi.r}\longrightarrow\mu$
and $\M\ni\xi$ for which
$\xi\not\stackrel{\mu.\xi.r}\longrightarrow\mu$. Theorem
\ref{thMLConv} can be generalized to i.i.d.\ sequences over
general finite alphabet $\X$.

The idea to prove $(ii)$ is to construct a sequence $x_{1:\infty}$
which is $\mu_{\th_0}\M$-random {\em and} $\mu_{\th_1}\M$-random
for $\th_0\neq\th_1$. This is possible if and only if $\Theta$
contains a gap and $\th_0$ and $\th_1$ are the boundaries of the
gap. Obviously $\xi$ cannot converge to $\th_0$ {\em and} $\th_1$,
thus proving $\M$-non-convergence. For no $\th\in[0,1]$ will this
$x_{1:\infty}$ be $\mu_\th$ M.L.-random. Finally, the proof of
Theorem \ref{thMLConv}
makes essential use of the mixture representation of $\xi$, as
opposed to the proof of Theorem \ref{thConv} which only needs
dominance $\xi\geqm\M$.

\section{Conclusions}\label{secConc}

For a hierarchy of four computability definitions, we completed
the classification of the existence of computable (semi)measures
dominating all computable (semi)measures. Dominance is an important
property of a prior, since it implies rapid convergence of the
corresponding posterior with probability one.
A strengthening would be convergence for all Martin-L{\"o}f (M.L.)
random sequences. This seems natural, since M.L.\ randomness can
be defined in terms of Solomonoff's prior $M$, so there is a close
connection.
Contrary to what was believed before, the question of posterior
convergence $M/\mu\to 1$ for all M.L.\ random sequences is still
open. We introduced a new flexible notion of $\mu/\xi$-randomness
which contains Martin-L{\"of} randomness as a special case. Though
this notion may have a wider range of application, the main
purpose for its introduction was to show that standard proof
attempts of $M/\mu\stackrel{M.L.}\longrightarrow 1$ based on
dominance only must fail. This follows from the
derived result that the validity of $\xi/\mu\to 1$ for
$\mu/\xi$-random sequences depends on the Bayes mixture $\xi$.


\end{document}